\documentclass[runningheads]{llncs}
\usepackage{graphicx}
\begin{document}
%
\title{TUDublin team at Constraint@AAAI2021 - COVID19 Fake News Detection}
%
%
\author{Elena Shushkevich\inst{1} \and
John Cardiff\inst{1}}
\authorrunning{E. Shushkevich et al.}
%
\institute{Technological University Dublin, Ireland\\
\email{elena.n.shushkevich@gmail.com}\\ 
\email{john.cardiff@tudublin.ie}\\
}
\maketitle              
\begin{abstract}
The paper is devoted to the participation of the TUDublin team in Constraint@AAAI2021 - COVID19 Fake News Detection Challenge. Today, the problem of fake news detection is more acute than ever in connection with the pandemic. The number of fake news is increasing rapidly and it is necessary to create AI tools that allow us to identify and prevent the spread of false information about COVID-19 urgently. The main goal of the work was to create a model that would carry out a binary classification of messages from social media as real or  fake news in the context of COVID-19. Our team constructed the ensemble consisting of Bidirectional Long Short Term Memory, Support Vector Machine, Logistic Regression, Naive Bayes and a combination of Logistic Regression and Naive Bayes. The model allowed us to achieve 0.94 F1-score, which is within 5\%  of the best result.  

\end{abstract}
\section{Introduction}
Nowadays, the increasing number of fake messages on social networks has become a huge problem. Social networks are the tools of communication that most of us use daily not only to communicate with friends and family, but also to get news about current events in the economy, politics, and the world around us in general.

Fake news is information hoaxes designed to deliberately mislead the reader in order to gain financial or political advantage\cite{ref_article1}. When faced with such news, it is not always easy to recognize whether the message is real or not, so it is necessary to create tools, including through the use of machine learning, that will help the unprepared reader to define if information they receive from social media is reliable or not. Such mechanisms should also help identify and prevent the spread of fake news by social media themselves.

In 2020, one of the main information agendas, without doubt, is the coronavirus, which has led to numerous losses, deaths, and economic difficulties. Often the main source of information on this topic for many people is social media, where news is updated promptly, as well as where it is possible to discuss what is happening in the world in social networks. Unfortunately, some people are trying to hype on such a global disaster as a pandemic, to get financial and political bonuses from what is happening. In this regard, there is fake news about the coronavirus, which spreads at a high speed. Such news can be not only false, but also dangerous for people's lives and health.

It is obvious that the detection and suppression of the spread of such fake news is an urgent goal, and in this article we are describing a possible method for detecting fake news about coronavirus, which we presented in the framework of the challenge Constraint@AAAI2021 - COVID19 Fake News Detection in English Language \cite{ref_article2}.

The article consists of the following 6 sections. Section 1 is the introduction to the problem. In Section 2 we conduct some relevant research in the field of identifying and investigating fake social media messages in the health sector and, in particular, fake news about the pandemic and COVID-19. In Section 3 we describe the challenge Constraint@AAAI2021 - COVID19 Fake News Detection and the dataset on which we trained and tested our models. In Section 4 we describe the steps we took during the preprocessing phase and the models we created to detect fake news about the Coronavirus. In Section 5 we present the results we achieved during the training phase and the results we achieved on the challenge and analyze them. The final Section 6 summarizes the conclusions that we have reached during the research and describes further possible improvements of the model.

\section{Related Work}

There are many works devoted to the study of fake news in the field of health care, such as \cite{ref_article3} where the authors' research questions were about ways that social media messages focused on fake health information or misinformation and health evidence with real or potential social impacts, and how interactions based on health evidence with social impacts help overcome fake health information. The authors used a methodology of social impact in social media, which combined quantitative and qualitative content analysis of the collected messages.

Numerous harmful advice about the prevention and treatment of diseases, taken seriously, could cause the deterioration of the patient's condition and the emergence of new diseases. In this regard, identifying fake news about the Coronavirus, as well as trying to prevent the further spread of such news, are important steps towards improving the current situation.
At the time of writing, although less than a year has passed since the emergence of COVID-19 as a global threat, already several studies have been published aimed at detecting and classifying fake messages in social media, related to COVID-19.

In one such work \cite{ref_article4} the authors researched the impact of rumors and misinformation in social media in a context of COVID-19 pandemic. They found that the huge number of hoaxes and misinformation leads to an increase in the spread of the virus and a decrease of mental health individuals. 

Researchers have used different Natural Language Processing and Machine Learning tools and models to solve these problems. In  \cite{ref_article5}, the authors collected 4,800 expert-annotated social media posts related to COVID-19 and 86 common misconceptions about the disease to evaluate the performance misinformation detection related to Coronavirus pandemic. 

For the modelling, the authors used TF-IDF\footnote[1]{\url{https://scikit-learn.org/stable/modules/generated/sklearn.feature\_extraction.text.\\TfidfVectorizer.html}} and GloVe\footnote[2]{\url{https://nlp.stanford.edu/projects/glove/}} to obtain vectorized representations, and also to obtain contextual word embeddings the RoBERTa-base \cite{ref_article6} implementation was used with two models of textual similarity. Also, they tuned RoBERTa-base using the dataset of tweets connected with COVID-19. 

In the research \cite{ref_article7} researchers used K-Nearest Neighbour Classifier to predict fake news connected with COVID-19.
The authors of \cite{ref_article8} used natural language processing to analyse the messages connected with COVID-19 and to indicate the main topics which people discuss in social media in the framework of the pandemic, where top negative themes included concerns about social distancing and isolation policies, misinformation, while top positive themes connected with the pandemic were public awareness, spiritual support, encouragement, charity and entertainment.

In addition to such studies, open challenges appear that invite participants to create effective models for identifying fake messages about the pandemic. The challenge we took part in is called Constraint@AAAI2021 - COVID19 Fake News Detection in English and Hindi, and we used the English dataset only.

\section{Dataset}

The goal of the Constraint@AAAI2021 - COVID19 Fake News Detection challenge was to create a model that would allow us to determine whether the message about COVID-19 is fake or real news.

the organizers of the challenge created an annotated dataset of 10,700 social media posts and articles of real and fake news on COVID-19 in English. To collect real tweets, they used relevant sources such as World Health Organization (WHO), Centers for Disease Control and Prevention (CDC), Covid India Seva, and others. To collect fake news, authors used Facebook posts, Twitter tweets, Instagram posts and other types of social media and checked them by fact-verification websites such as PolitFact\footnote[3]{\url{https://www.politifact.com/}} and Boomlive\footnote[4]{{https://www.boomlive.in/}} to ensure the messages were fake. The collected dataset was split into train (60\%), validation (20\%), test (20\%).

As a result, authors collected 6,420 training messages (3,360 of them were real and 3,060 - fake ones), 2,140 validation messages (1,120 - real, 1,020 - fake), and 2,140 testing messages (with 1,120 real and 1,010 fake messages).

\section{Model}

This section describes the two stages of our model creation: the preprocessing and the modelling steps.

\subsection{Preprocessing}

Preprocessing is an important step in creating a model that allows researchers to identify fake news in social media, because it helps to prepare the data for subsequent processing.
In our case, at the preprocessing stage we:

- used TF-IDF (term frequency - inverse document frequency) for the classical machine learning models;

- used PorterStemmer\footnote[5]{\url{https://www.nltk.org/howto/stem.html}} which removes morphological affixes from words, leaving only the word stem, for the neural networks models;

- translated all the letters of the dataset to lowercase.

After performing preprocessing, we started the modeling stage.

\subsection{Modelling}

At the modelling stage, we used both classical machine learning models and models based on neural networks. Also, we implemented ensembles that allow us to combine the results of several different models with the aim to achieve the best results.

In the process of modeling based on classical machine learning, we implemented the following models:

- Logistic regression (LR)\cite{ref_article9,ref_article10} one of the most popular approaches with the basic idea of a linear classifier is that a feature space can be divided by a hyperplane into two half-spaces, in each of which one of the two values of the target class is predicted.

- SVM \cite{ref_article11,ref_article12,ref_article13} which creates a hyperplane or set of hyperplanes in multidimensional or infinite-dimensional space that can be used to solve classification, regression, and other related problems.

- Naive Bayes (NB)\cite{ref_article14} which is quite simple and fast to use and is used as a reference point when comparing different methods. Also, Naive Bayes method needs a priori information about probabilities of each class, while other methods don't need it.

- NB+LR - the interpolation between LR and NB with the coefficient of interpolation as a form of the regularization: in practice it means that in this type of modeling we trust NB unless the LR is very confident, which is the analogy of the approach from \cite{ref_article15} and where such type of algorithms overperformed another approaches.

Having built these models, we combined them into an ensemble that resulted in the final class of the message (fake or real news): each model in the ensemble showed the probability of class,  we the summed up such probability and divided the sums to the number of models, then we compared the obtained values and chose the largest (and, accordingly, the class). The approach based on the ensembles constructing has proven itself well in classification problems \cite{ref_article16,ref_article17,ref_article18,ref_article19}. If the final arithmetic mean probability of belonging to the class of fake news was higher than the final arithmetic mean probability of belonging to the class of real news, then we marked the message as fake news.

Additionally, we have created models for detecting fake news based on neural networks, namely:
- LSTM (Long Short-Term Memory) \cite{ref_article20,ref_article21} with an embedding layer, which is a specific kind of recurrent neural network, and is capable of learning long-term dependencies. 
- Bidirectional LSTM with an embedding layer, which consists of two LSTMs that are run in parallel, and which achieved good results in \cite{ref_article22}.

\begin{figure}
\includegraphics[scale=0.50] {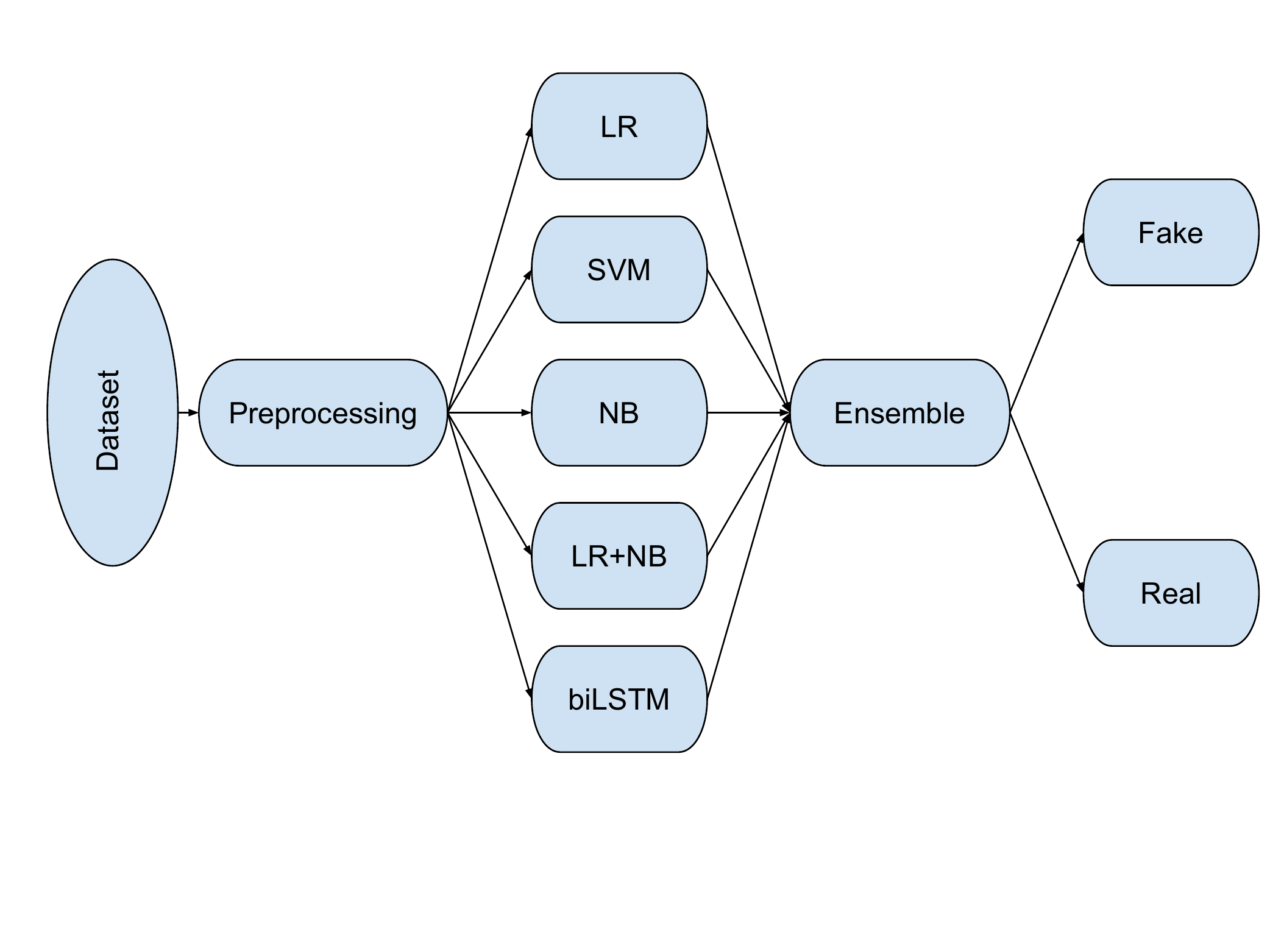}
\caption{The construction of final ensemble of models.} \label{fig1}
\end{figure}

Finally, we have combined into a single ensemble the models of classical machine learning and bidirectional LSTM, the structure of which is shown in Figure \ref{fig1}.

The final class of the messages was determined by the majority voted method.

\section{Results}

The results we achieved on the training dataset are presented in Table \ref{tab1}.

\begin{table}
\caption{Results with Classical Machine Learning Models on training dataset.}\label{tab1}
\begin{center}
\begin{tabular}{r@{\quad}rl}
\hline
\multicolumn{1}{l}{\rule{0pt}{12pt}
                   Model}&\multicolumn{2}{l}{F1-score}\\[2pt]
\hline\rule{0pt}{12pt}
Logistic Regression (LR)  &     0.92 \\
Support Vectors Machine (SVM)  &   0.91 \\
Naive Bayes & 0.93\\
NB+LR & 0.93\\
Ensemble & 0.93\\[2pt]
\hline
\end{tabular}
\end{center}
\end{table}

As we can see from the table, all models show similar fairly high results, because of which we decided to include an ensemble in the final submission, since it combines all models and we decided to check our own ensemble on the testing dataset.

The results of neural network models we obtained on the training dataset are presented in Table \ref{tab2}.

\begin{table}
\caption{Results with Neural Networks Models on training dataset.}\label{tab2}
\begin{center}
\begin{tabular}{r@{\quad}rl}
\hline
\multicolumn{1}{l}{\rule{0pt}{12pt}
                   Model}&\multicolumn{2}{l}{F1-score}\\[2pt]
\hline\rule{0pt}{12pt}
Long Short-Term Memory (LSTM)  &     0.94 \\
Bidirectional Long Short-Term Memory (biLSTM)  &   0.95 \\[2pt]
\hline
\end{tabular}
\end{center}
\end{table}

Both models showed decent results, so we included them in the final submission.

Table \ref{tab3} shows the results we achieved with the all different models we submitted.

\begin{table}
\caption{Results on testing dataset.}\label{tab3}
\begin{center}
\begin{tabular}{r@{\quad}rl}
\hline
\multicolumn{1}{l}{\rule{0pt}{12pt}
                   Model}&\multicolumn{2}{l}{F1-score}\\[2pt]
\hline\rule{0pt}{12pt}
SVM + LR+Naive Bayes + (LR+NB) + biLSTM  &     0.94 \\
SVM + Naive Bayes + (LR+NB)  &   0.94 \\
LSTM & 0.92\\
biLSTM & 0.91\\[2pt]
\hline
\end{tabular}
\end{center}
\end{table}

The results we obtained indicate that in our case both ensembles of models work better than the single models. Note that the best result was achieved with an ensemble that combines both classical machine learning models and a neural network, which indicates the prospects for the future work in this direction. Also, it should be noted that at the training step results of biLSTM overperformed LSTM, while on the testing dataset LSTM model showed better result. In future we are planning to investigate that case and explain this difference.
 All the results obtained are quite close, within 4\% of each other.

Considering the results we achieved in the context of the challenge, it should be noted that the best F1 score is 0.99. This result differs from our best result by less than 5\%, which indicates the competitiveness of the model we have built.

\section{Conclusion}

To sum up, it should be noted that the problem of fake news detection and in particular fake news about COVID-19, is an important and urgent issue. The challenge Constraint@AAAI2021-COVID19 Fake News Detection, which aimed at detecting fake news about COVID-19 in social networks, could help to create models that can quickly identify and stop the spreading of this kind of fake news.

The TUDublin team created several models based on both classical machine learning models and models based on neural networks, as well as ensembles that combine several of these models. The best results were achieved using ensembles, and these results are competitive compared to the results of the challenge winners. Also, it should be noted that our study was more experimental, this way we had aim to check our models on the testing dataset.

In the future, we are planning to further develop our work, paying more attention to the creation and training of neural networks at a more advanced level, since, as we have noted above, neural networks allow us to identify and classify fake messages with high effectiveness. We are also planning to pay more attention to linguistic features that would improve the quality of classification, and include additional datasets so that neural networks are trained on more data.

\bibliographystyle{splncs04}
%

\end{document}